\pdfoutput=1

\documentclass[11pt,dvipsnames]{article}

\usepackage[]{acl}

\usepackage{times}
\usepackage{latexsym}
\usepackage{multicol} 
\usepackage{graphicx,calc}
\newlength\myheight
\newlength\mydepth
\settototalheight\myheight{Xygp}
\usepackage[T1]{fontenc}

\usepackage[utf8]{inputenc}

\usepackage{microtype}
\usepackage[algo2e]{algorithm2e}
\usepackage{times}
\usepackage{latexsym}
\usepackage{amsmath}

\usepackage{multirow}
\usepackage{makecell}

\usepackage[T1]{fontenc}

\usepackage[utf8]{inputenc}

\usepackage{microtype}
\usepackage{booktabs}
\usepackage{graphicx}
\usepackage{algorithm}
\usepackage{algorithmicx}
\usepackage{amsmath}
\usepackage{caption}
\usepackage{subfigure}
\usepackage{bbding}
\usepackage{color, soul, colortbl}
\usepackage[most]{tcolorbox}
\usepackage{lipsum} 

\definecolor{myred}{rgb}{0.839,0.341,0.271}
\definecolor{myblue}{rgb}{0.843,0.898,0.941}
\definecolor{mypurple}{rgb}{0.957,0.937,0.953}

\definecolor{myerror}{rgb}{0.996, 0.875,0.863}
\usepackage{tikz}
\usepackage{makecell}
\usepackage{cleveref}
\usepackage{ragged2e}
\crefname{section}{§}{§§}
\Crefname{section}{§}{§§}
\usepackage{tikz,lipsum}
\usepackage[most]{tcolorbox}
\usepackage{awesomebox} 
\usepackage[tikz]{bclogo}
\usepackage[framemethod=tikz]{mdframed}
\definecolor{bgblue}{RGB}{245,243,253}
\definecolor{ttblue}{RGB}{91,194,224}
\usepackage{enumitem}

\title{Enhancing Chat Language Models by Scaling High-quality \\ Instructional Conversations}

\author{Ning Ding, Yulin Chen, Bokai Xu, Yujia Qin,  \textbf{Zhi Zheng}, \textbf{Shengding Hu} \\  \textbf{Zhiyuan Liu}, \textbf{Maosong Sun},  \textbf{Bowen Zhou}\\
 Tsinghua University\\ 
}
\begin{document}
\maketitle
\begin{abstract}

Fine-tuning on instruction data has been widely validated as an effective practice for implementing chat language models like ChatGPT. Scaling the diversity and quality of such data, although straightforward, stands a great chance of leading to improved performance.
This paper aims to improve the upper bound of open-source models further. 
We first provide a systematically designed, diverse, informative, large-scale dataset of instructional conversations, UltraChat, which does not involve human queries.
Our objective is to capture the breadth of interactions that a human might have with an AI assistant and employs a comprehensive framework to generate multi-turn conversation iteratively.
UltraChat contains 1.5 million high-quality multi-turn dialogues and covers a wide range of topics and instructions. 
Our statistical analysis of UltraChat reveals its superiority in various key metrics, including scale, average length, diversity, coherence, etc., solidifying its position as a leading open-source dataset.
Building upon UltraChat, we fine-tune a LLaMA model to create a powerful conversational model, UltraLLaMA.
Our evaluations indicate that UltraLLaMA consistently outperforms other open-source models, including Vicuna, the previously recognized state-of-the-art open-source model.
The dataset and the model will be publicly released\footnote{\url{https://github.com/thunlp/UltraChat}}.

\end{abstract}

\section{Introduction}

\begin{table}[!t]
    \centering
    \scalebox{0.88}{
    \begin{tabular}{lc} \toprule
        \textbf{Model} &  \textbf{Score} \\ \midrule
        {\includegraphics[height=1.2\myheight]{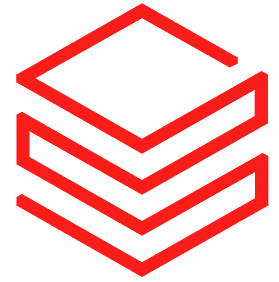}} Dolly-v2~\cite{Mike2023dolly} & 7.145 ± 2.773 \\
        {\includegraphics[height=1.2\myheight]{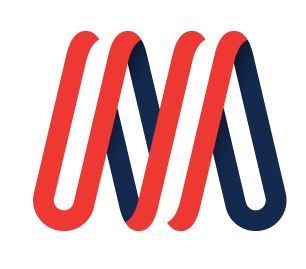}} MPT-Chat~\cite{mpt}  &  8.317 ± 2.117\\
         {\includegraphics[height=1.2\myheight]{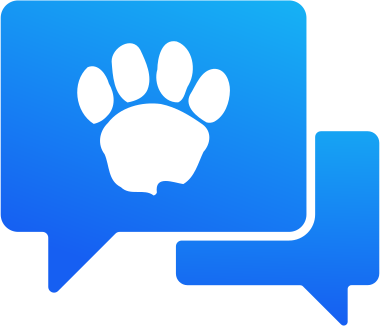}} OpenAssistant~\cite{kopf2023openassistant}  &  8.470 ± 1.505\\
        \raisebox{-\mydepth}{\includegraphics[height=1.3\myheight]{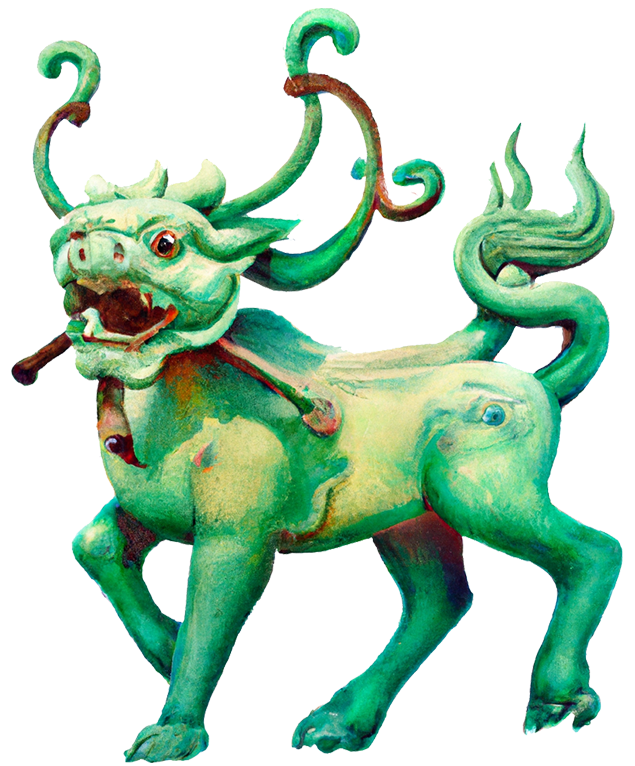}} Baize~\cite{xu2023baize} & 8.566 ± 0.986 \\
         \raisebox{-\mydepth}{\includegraphics[height=1.3\myheight]{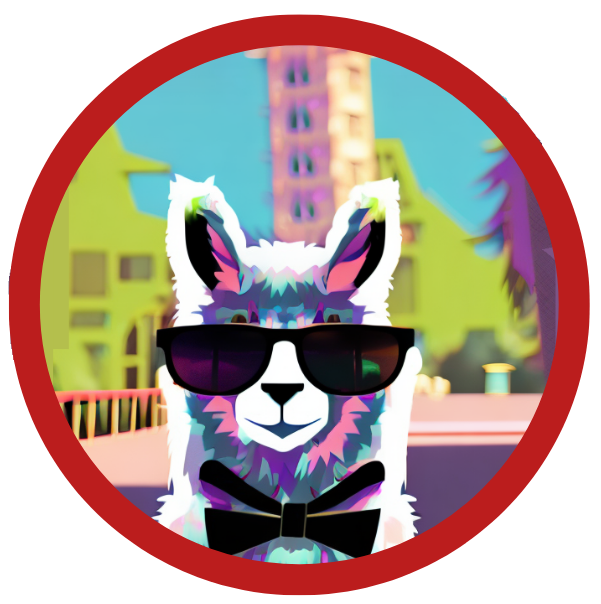}} {Alpaca}~\cite{taori2023alpaca} & 8.597 ± 1.292 \\
         \raisebox{-\mydepth}{\includegraphics[height=1.3\myheight]{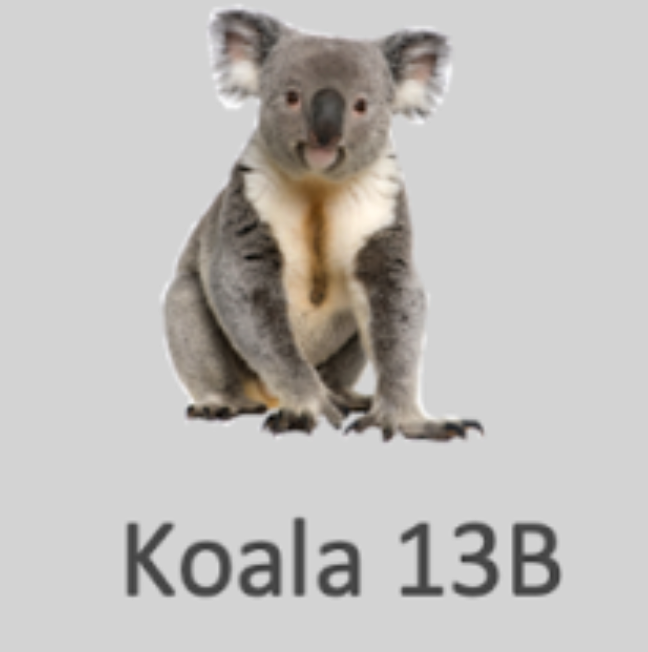}} Koala~\cite{koala_blogpost_2023} & 8.881 ± 1.062 \\
         \raisebox{-\mydepth}{\includegraphics[height=1.3\myheight]{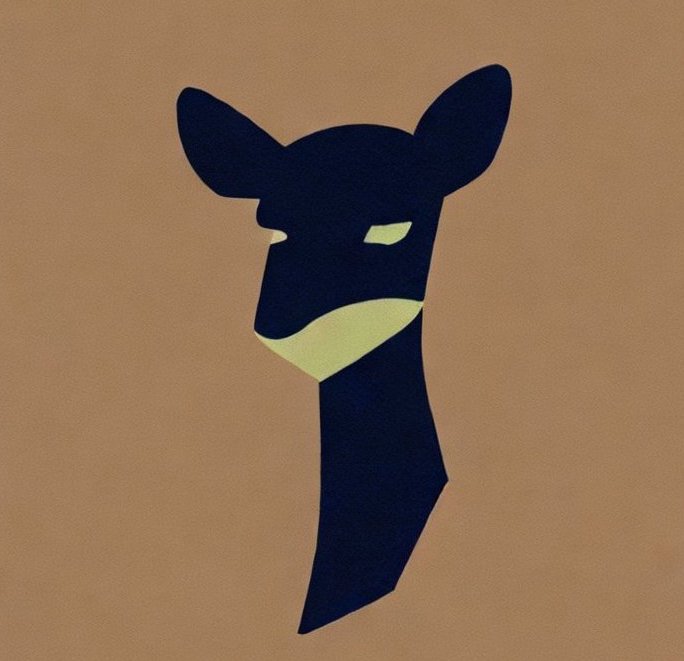}} Vicuna~\cite{vicuna2023} & 8.961 ± 0.718 \\ \midrule
         UltraLLaMA (ours) & \textbf{9.023 ± 0.952} \\ \bottomrule
    \end{tabular}}
    \caption{Average scores (1-10) across different open-source models and our model trained on UltraChat. The scores are independently assessed by ChatGPT, using a dataset consisting of over 300 questions generated by GPT-4. Among the models included in the evaluation, Dolly-v2 and OpenAssistant have 12B parameters, MPT-Chat and Alpaca have 7B parameters, while the remaining models have 13B parameters. 
    Evaluation prompts can be found in Appendix~\ref{sec:prompts}.}
    \label{tab:scores}
\end{table}

\looseness=-1 Large language models~\cite{HAN2021,bommasani2021opportunities} (LLMs) have demonstrated exceptional generalization capability on a variety of language-related tasks. 
Notably, ChatGPT~\cite{openai2022chatgpt}, an optimized version of GPT-3~\cite{brown2020language} and GPT-4~\cite{openai2023gpt} for conversation, takes the user experience to another level via excelling in comprehending and generating responses in a natural and interactive manner. 
The introduction of ChatGPT has spurred a surge in the adoption and implementation of general chat models.


In addition to competing models developed by large corporations such as Bard\footnote{\url{https://bard.google.com/}} and Claude\footnote{\url{https://www.anthropic.com/index/introducing-claude}}, the open-source community is actively engaged in training similar models, aiming to democratize access to AI technology. Notable examples in this regard include Alpaca~\cite{taori2023alpaca}, Vicuna~\cite{vicuna2023}, Koala~\cite{koala_blogpost_2023}, Baize~\cite{xu2023baize}, and Belle~\cite{belle2023exploring}, etc., demonstrating promising performance. 
Experimental evidence strongly suggests that chat language models can be effectively trained through instruction fine-tuning~\cite{wei2021finetuned,sanh2021multitask}, and they also indicate that many data-efficient~\cite{Zhou2023LIMALI} or computing-efficient~\cite{hu2021lora,ding2023parameter} methods can be applied.
This paper, in another way, focuses more on the "final one mile" of chat language models, as evidence shows that \textbf{the journey from 0 to 60 is easy, whereas progressing from 60 to 100 becomes exceedingly challenging}. 
For instance, researchers have shown that by utilizing a small, thoughtfully curated set of instructions, it is possible to train a model with satisfactory instruction-following capabilities. However, these approaches have yet to produce models that surpass the performance of Vicuna, the current leading open-source model, let alone outperform ChatGPT and GPT4.

This paper believes that the most straightforward way, that is, the quality and diversity of data employed in the training process, play a vital role in further improving the performance of chat language models. 
In other words, leveraging higher quality and more diverse data can yield better outcomes.
To this end, we present UltraChat, a million-scale multi-turn instructional conversation data to facilitate the construction of more powerful chat language models.
UltraChat is first designed by a principle that attempts to capture the breadth of interactions that a human might have with an AI assistant. 
Specifically, we do not use specific tasks like question-answering or summarization to construct the data, but curate three sectors: Questions about the World, Creation and Generation, and Assistance on Existing Materials. Then we employ meta-information, in-context expansion, and iterative prompting to scale up the number of instructions. To construct informative and realistic multi-turn conversations, two separate ChatGPT Turbo APIs are adopted in the conversation generation, where one plays the role of the user to generate queries, and the other generates the response. We instruct the user model with carefully designed prompts to mimic human user behavior and call the two APIs iteratively. 

We fine-tune a LLaMA-13B model on UltraChat to produce UltraLLaMA and compare the model to a wide range of baselines, especially the open-source ones. The evaluation shows that our model could consistently outperform other models. As reported in Table~\ref{tab:scores}, UltraLLaMA can achieve the highest performance scores that are independently assessed by ChatGPT. We also perform a preference study to make ChatGPT choose a response with higher overall performance, the results show that our model still consistently outperforms all the open-source baselines.





\section{Related Work}

\paragraph{Instruction Tuning.}
Recent works demonstrate LLMs with powerful capabilities in following human instructions. \citet{wei2021finetuned} pioneered to fine-tune T5~\citep{2020t5} on $60$ NLP datasets verbalized with natural language instruction templates, i.e., \textit{instruction tuning}. The fine-tuned model exhibits a strong ability in instruction understanding and generalizes well to unseen instructions (tasks). Later, \citet{longpre2023flan} extend the setting to $1,836$ tasks and show the benefits of scaling the number of tasks in out-of-distribution generalization. \citet{wei2021finetuned} also conclude that the success of instruction tuning depends on the quality of the dataset and the design of prompts. To further regulate the tuned model's behavior, \citet{ouyang2022training,schulman2017proximal} propose to first learn a reward model directly from annotated human feedback, then employ reinforcement learning to align model behaviors with human preferences. This technique can be combined with instruction tuning to further boost the model performance and has been successfully applied to LLMs such as ChatGPT.

\begin{figure*}[!ht]
    \centering
    \includegraphics[width=\linewidth]{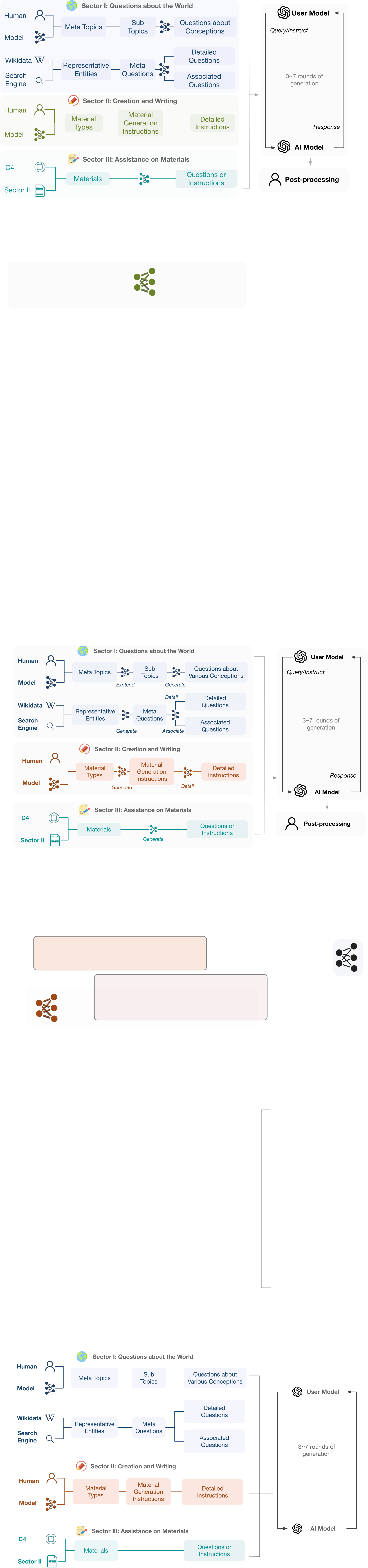}
    \caption{Construction process of UltraChat. The three sectors of data are derived from different meta information.}
    \label{fig:ulrta}
\end{figure*}

\paragraph{Data Augmentation with LLMs.}
Collecting large-scale human-annotated instructions and their responses is time-consuming and labor-intensive. Alternatively, a more cost-effective and feasible approach to gathering top-notch data involves sampling from LLMs that have been finely tuned, e.g., ChatGPT and GPT-3.5. 
Recently, there is a surge of interest in distilling these powerful LLMs for data augmentation. For instance, using the technique of SelfInstruct~\citep{wang2022selfinstruct}, Alpaca~\citep{alpaca} generate $52$k high-quality instruction-response pairs based on $175$ seed tasks by ``distilling'' Text-Davinci-003. After training a LLaMA~\citep{touvron2023llama} model on the dataset, the model performs almost on par with Text-Davinci-003. The success of Alpaca boosts numerous later efforts on data augmentation with LLMs, such as code-alpaca~\citep{codealpaca}, alpaca-cot~\citep{alpaca-cot}, GPT4ALL~\citep{gpt4all}, ShareGPT~\citep{domeccleston2023sharegpt}, Dolly-v2~\citep{Mike2023dolly}, BELLE~\citep{belle2023exploring}, Vicuna~\citep{vicuna2023}, Koala~\citep{koala_blogpost_2023}, Baize \cite{xu2023baize}, etc. It is shown that increasing the scale of data could constantly improve the model performance. Besides scaling the data size, these works also diverge in their ways of prompt engineering to gather data with better quality. For instance, CAMEL~\citep{li2023camel} designed a multi-agent role-play environment for LLMs to solve a given complex task and produced $115$k instruction-response pairs that simulate real human conversations.

\section{Design}

\looseness=-1 LLMs are believed to be better annotators than human-being in many scenarios~\cite{gilardi2023chatgpt}. 
However, directly using LLMs like ChatGPT to generate multi-turn conversations can be satisfactory but not informative, as it cannot enjoy the benefit of reinforcement learning with human feedback (RLHF) in the alignment process. 
Table~\ref{tab:compare} in Appendix~\ref{sec:example} shows a comparison of directly generated multi-turn dialogue and a case in UltraChat with the same opening line.
Two key points can be derived to ensure the quality of the data: (1) An \textit{opening line} directly determines the topic of the dialogue.  Opening lines should be highly diverse and encompass any task that a human user may request a chat model to perform.
(2) A \textit{user} determines the plot of the dialogue, and the output should be tailored to the current topic with diverse language styles and requests.

 UltraChat aims to cover a tremendous range of instructions and queries, which is composed of three sectors: \textit{Questions about the World}, \textit{Creation and Generation}, and \textit{Assistance on Existing Materials}.

\subsection{Principle}

Unlike other datasets that tend to use specific tasks, such as question-answering, rewriting, and summarization, to construct the data, the design of our schema is grounded in a tripartite framework that aims to capture the breadth of interactions that a human might have with an AI assistant. We believe any interactions between a human user and an AI assistant can be regarded as obtaining information.


\paragraph{Information Access.} The first sector, "Questions about the world," focuses on querying existing information in the world.
This is a key aspect of human-AI interaction, as users often rely on AI assistants to provide quick and accurate answers to their questions. By including a wide range of topics, the dataset addresses the diverse information needs of users, ensuring that the AI assistant can provide relevant and comprehensive responses. This component is crucial in facilitating effective information exchange between the user and the AI assistant, which is at the core of any human-AI interaction.

\paragraph{Conditional Information Creation.} The second part, "Creation and writing," is concerned with the creation of new information with human-input 
 conditions. This process reflects the AI's capacity to engage in creative tasks alongside users, harnessing its vast knowledge and pattern recognition capabilities to generate original content. This part of the dataset acknowledges the role of AI assistants as collaborative partners in the creative process, pushing the boundaries of what AI can achieve and enabling users to harness its potential for a wide range of tasks, from writing emails to crafting stories and plays.

\paragraph{Information Transformation.} The third part, "Assistance on Existing Materials," addresses the modification of existing information. This is a crucial aspect of human-AI interaction, as it allows the AI assistant to actively engage with the user's input, transforming it in various ways, such as through rewriting, continuation, summarization, or inference. This part of the dataset ensures that the AI assistant is capable of manipulating information to better serve the user's needs, enabling it to function as a versatile and adaptive tool that can handle a diverse array of tasks.

In summary, this tripartite principle is designed to provide a comprehensive representation of the possible interactions between humans and AI assistants. 
Based on the design, we create UltraChat to capture meaningful and efficient collaboration between human users and AI systems.. 




\begin{table*}[!ht]
    \centering
    \scalebox{0.86}{
    \begin{tabular}{lll} \toprule
        \raisebox{-\mydepth}{\includegraphics[height=1.3\myheight]{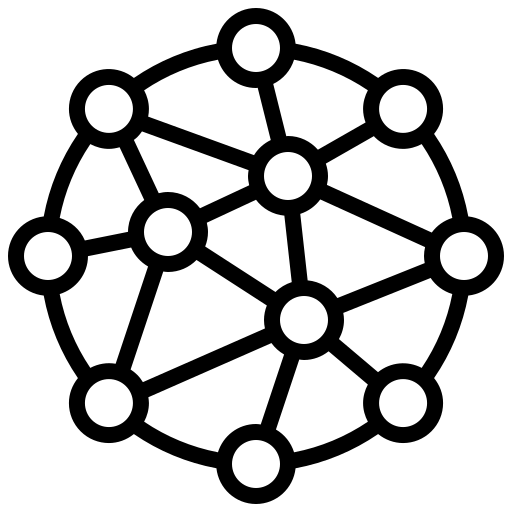}} Technology  &  \raisebox{-\mydepth}{\includegraphics[height=1.3\myheight]{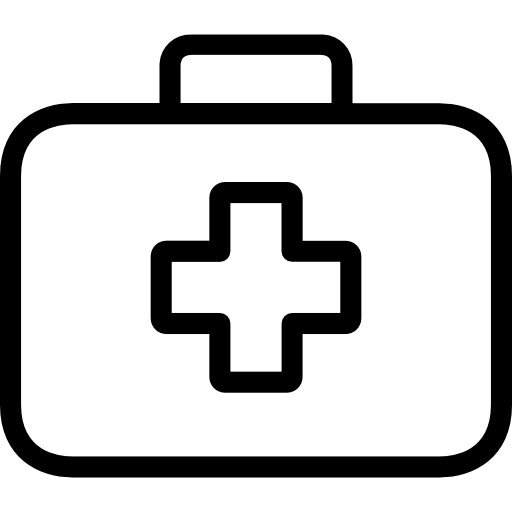}} Health and wellness &  \raisebox{-\mydepth}{\includegraphics[height=1.3\myheight]{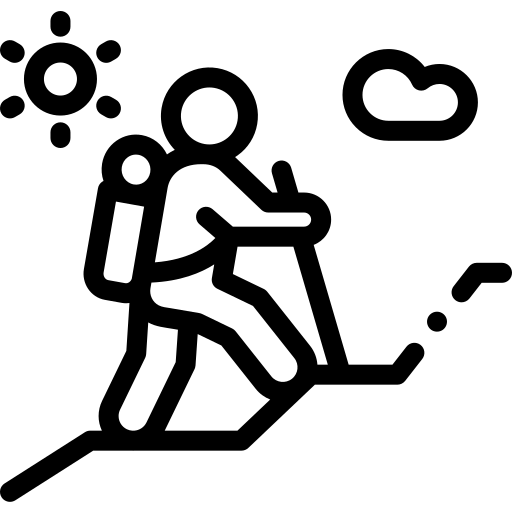}} Travel and adventure \\ 
        \raisebox{-\mydepth}{\includegraphics[height=1.3\myheight]{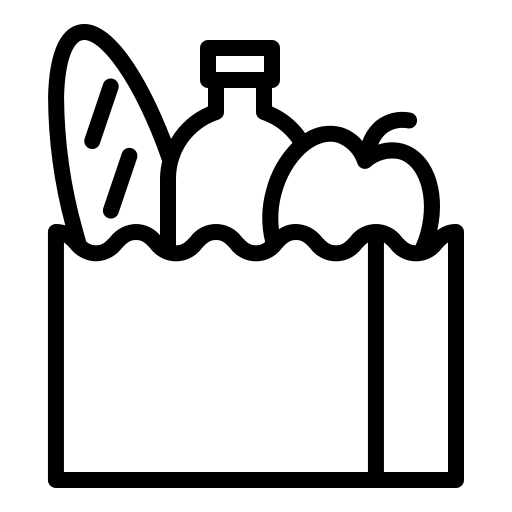}} Food and drink &  \raisebox{-\mydepth}{\includegraphics[height=1.3\myheight]{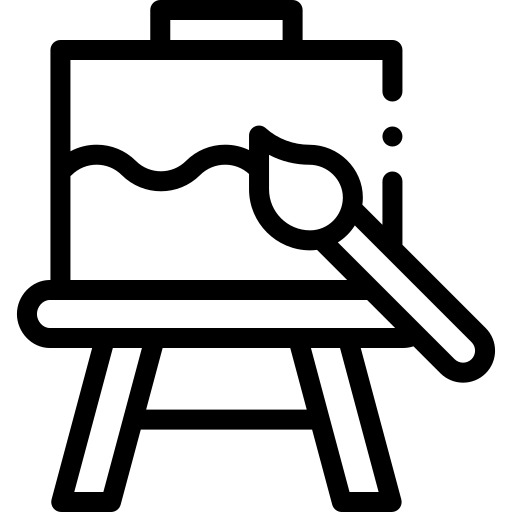}} Art and culture &  \raisebox{-\mydepth}{\includegraphics[height=1.3\myheight]{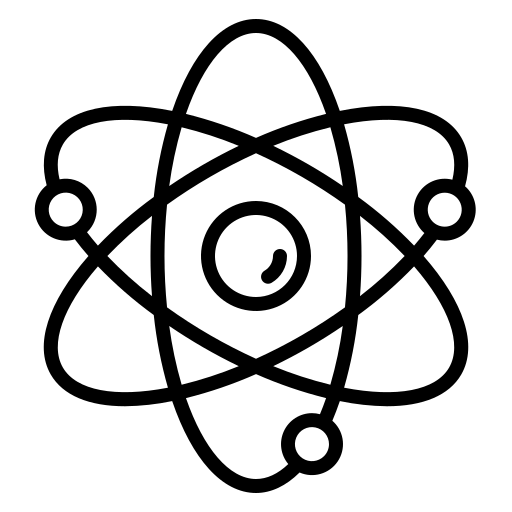}} Science and innovation  \\
        \raisebox{-\mydepth}{\includegraphics[height=1.3\myheight]{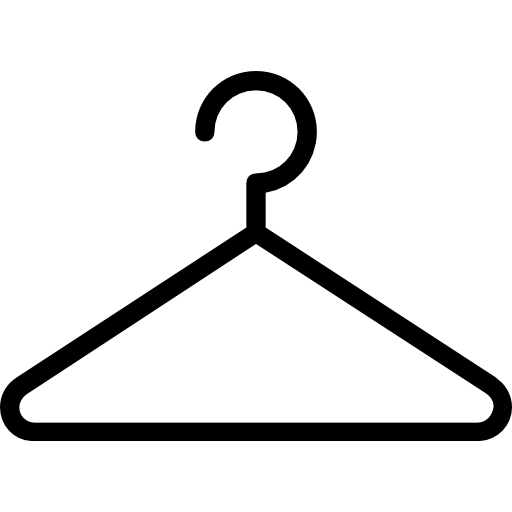}} Fashion and style & 
  \raisebox{-\mydepth}{\includegraphics[height=1.3\myheight]{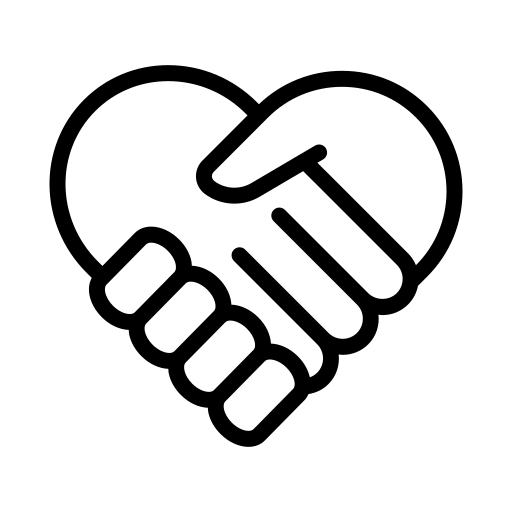}} Relationships and dating & {\includegraphics[height=1.3\myheight]{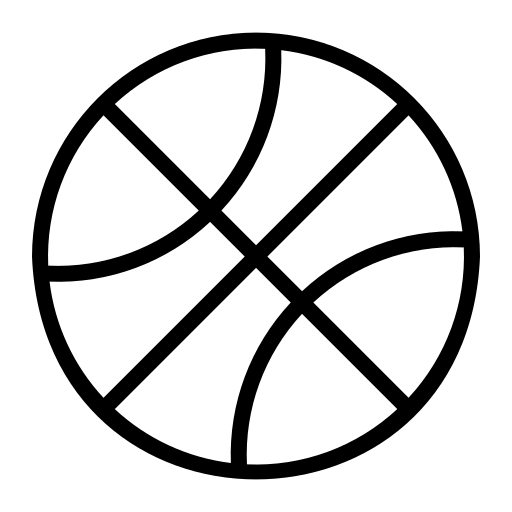}} Sports and fitness \\
        \raisebox{-\mydepth}{\includegraphics[height=1.3\myheight]{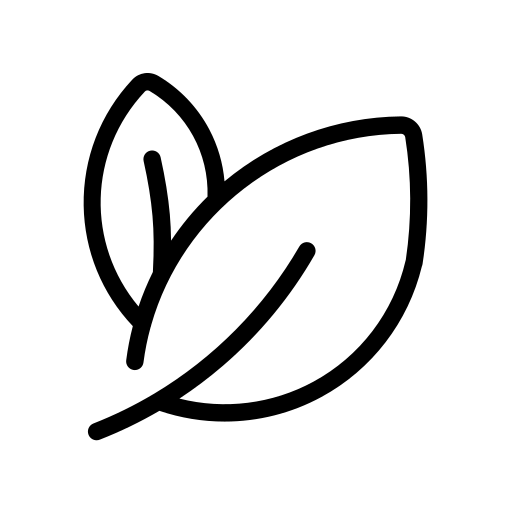}} Nature and the environment & 
         \raisebox{-\mydepth}{\includegraphics[height=1.3\myheight]{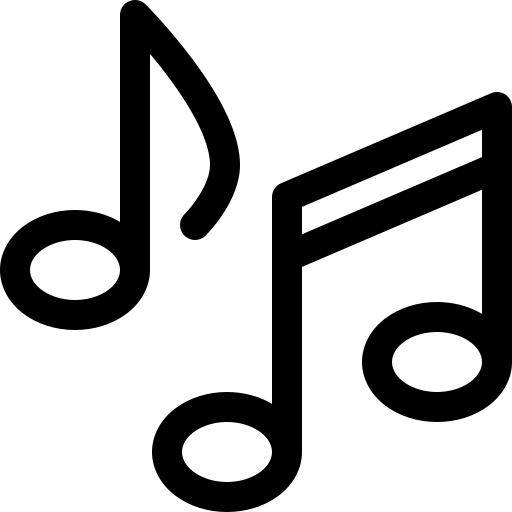}} Music and entertainment &  \raisebox{-\mydepth}{\includegraphics[height=1.3\myheight]{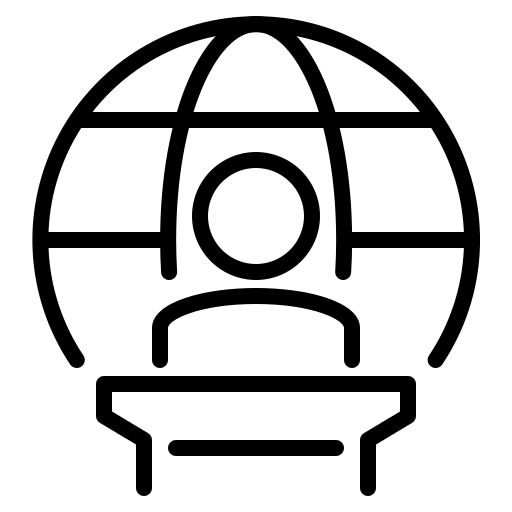}} Politics and current events \\
        \raisebox{-\mydepth}{\includegraphics[height=1.3\myheight]{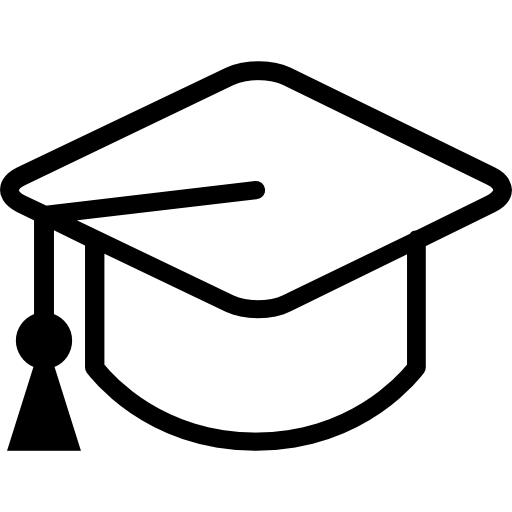}} Education and learning &   \raisebox{-\mydepth}{\includegraphics[height=1.3\myheight]{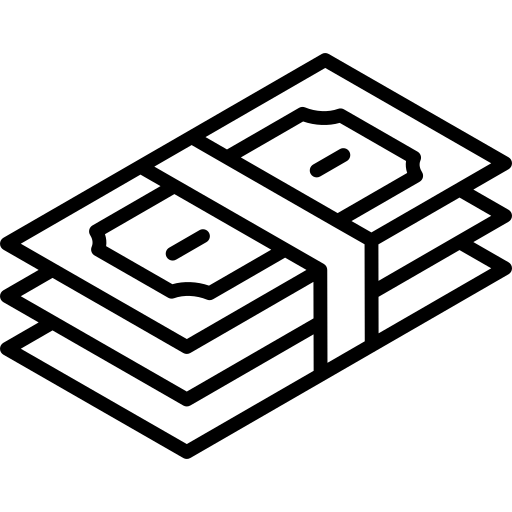}} Money and finance &  \raisebox{-\mydepth}{\includegraphics[height=1.3\myheight]{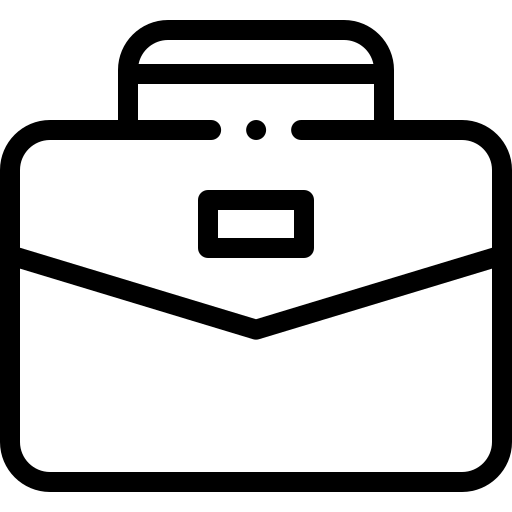}} Work and career \\
        \raisebox{-\mydepth}{\includegraphics[height=1.3\myheight]{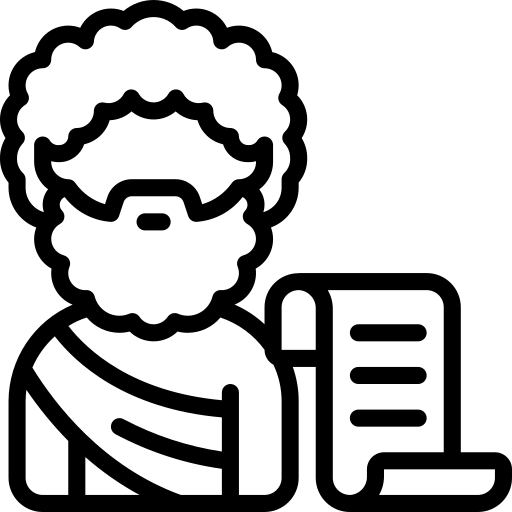}} Philosophy and ethics &  \raisebox{-\mydepth}{\includegraphics[height=1.3\myheight]{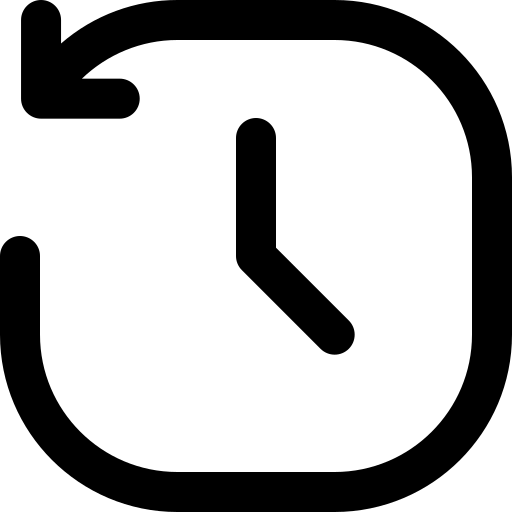}} History and nostalgia &
        \raisebox{-\mydepth}{\includegraphics[height=1.3\myheight]{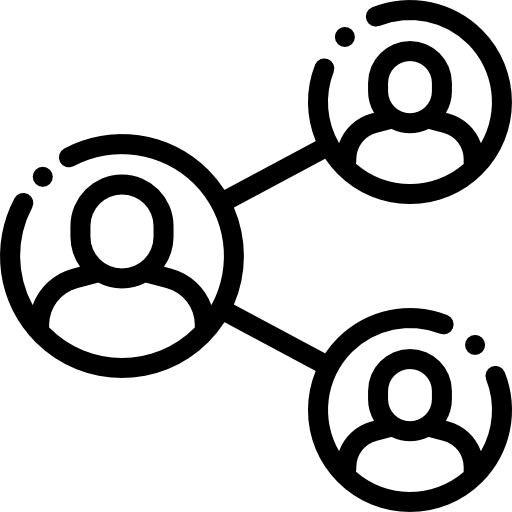}} Social media and communication \\
         \raisebox{-\mydepth}{\includegraphics[height=1.3\myheight]{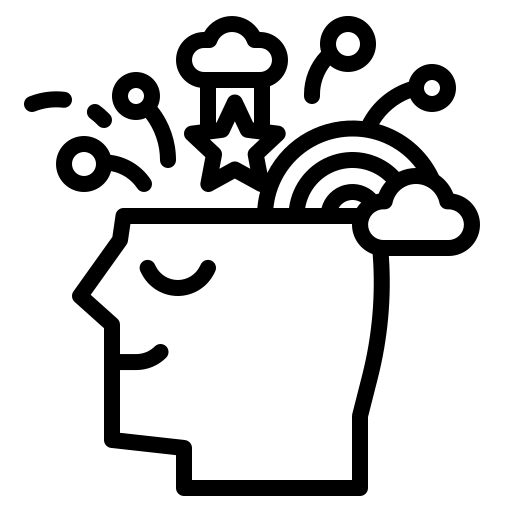}} Creativity and inspiration & 
         {\includegraphics[height=1.3\myheight]{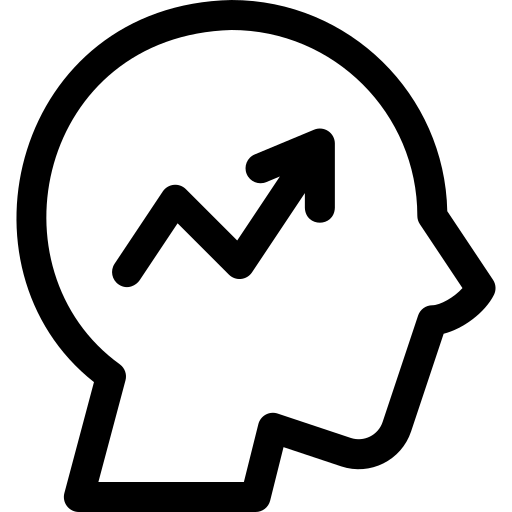}} Personal growth and development &  \raisebox{-\mydepth}{\includegraphics[height=1.3\myheight]{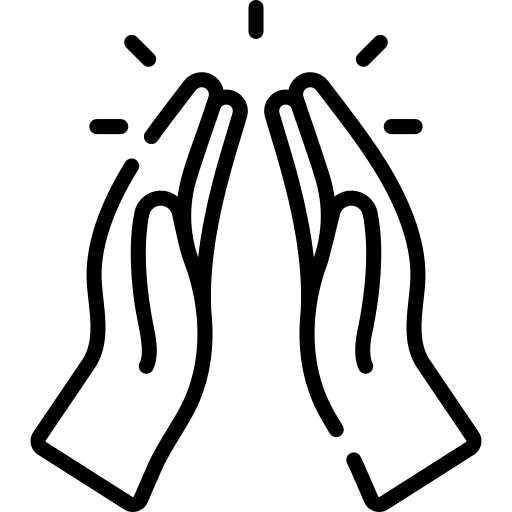}} Spirituality and faith \\
         \raisebox{-\mydepth}{\includegraphics[height=1.3\myheight]{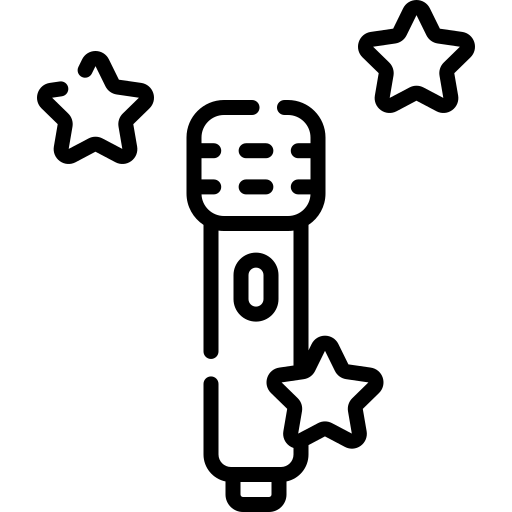}} Pop culture and trends &   \raisebox{-\mydepth}{\includegraphics[height=1.3\myheight]{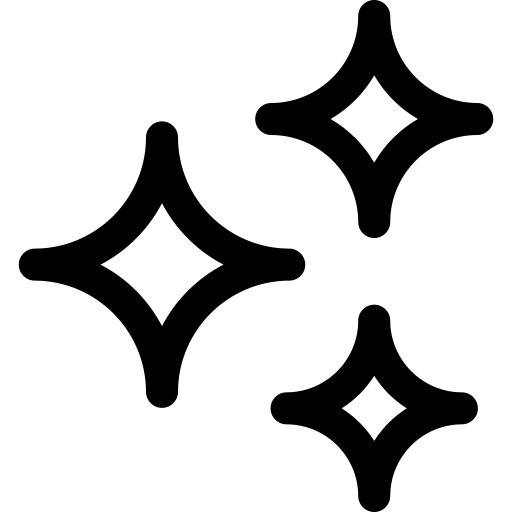}} Beauty and self-care &  \raisebox{-\mydepth}{\includegraphics[height=1.3\myheight]{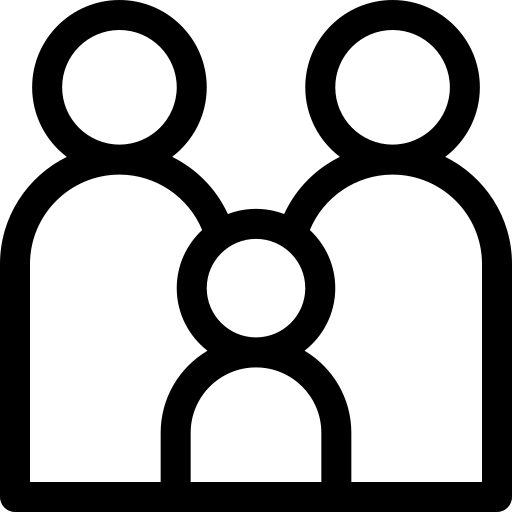}} Family and parenting \\
         \raisebox{-\mydepth}{\includegraphics[height=1.3\myheight]{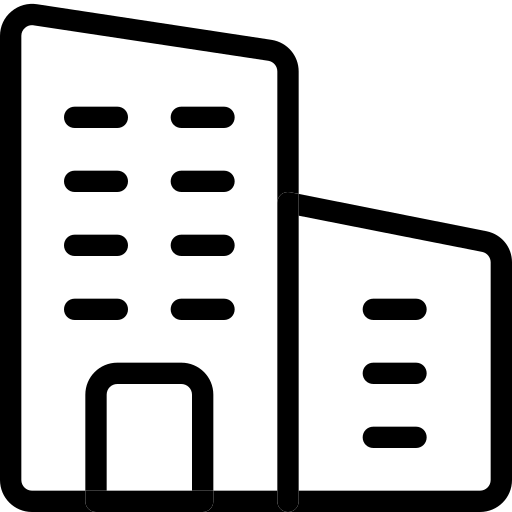}} Entrepreneurship and business & {\includegraphics[height=1.3\myheight]{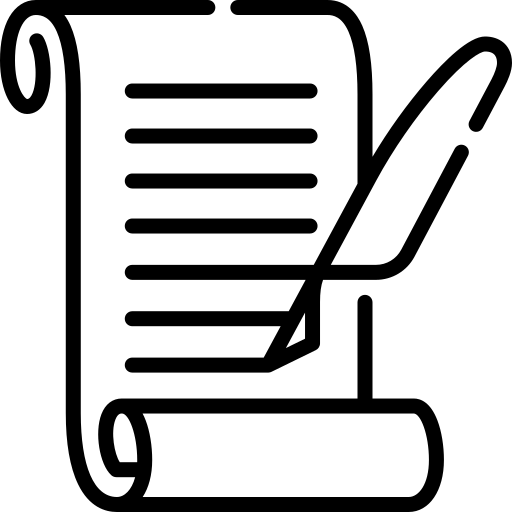}} Literature and writing &  \raisebox{-\mydepth}{\includegraphics[height=1.3\myheight]{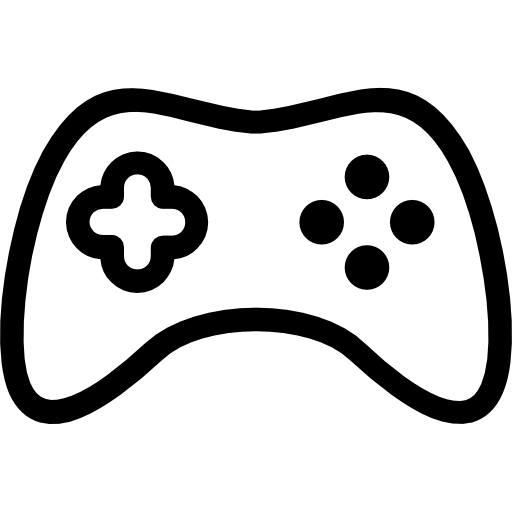}} Gaming and technology \\ 
         \raisebox{-\mydepth}{\includegraphics[height=1.3\myheight]{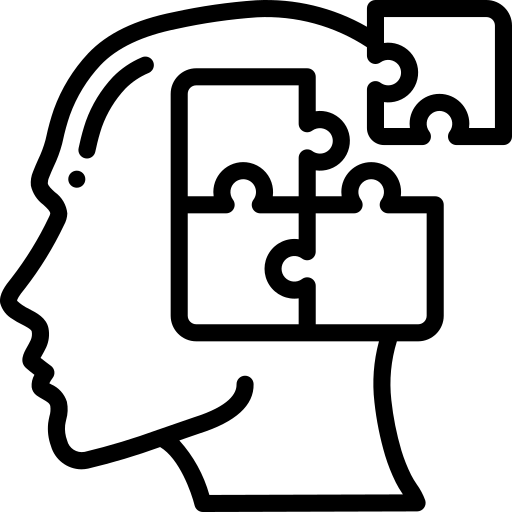}} Mindfulness and meditation & {\includegraphics[height=1.3\myheight]{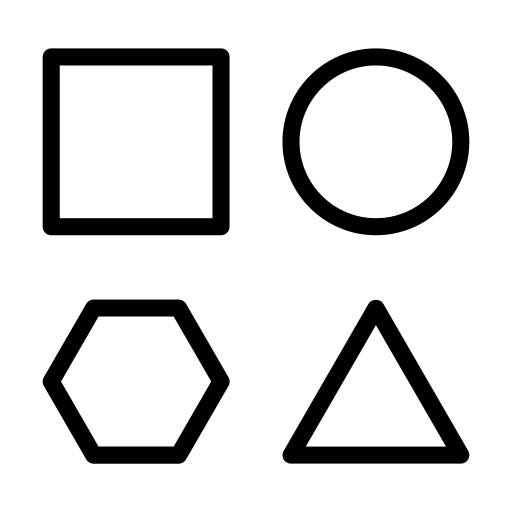}} Diversity and inclusion &  \raisebox{-\mydepth}{\includegraphics[height=1.3\myheight]{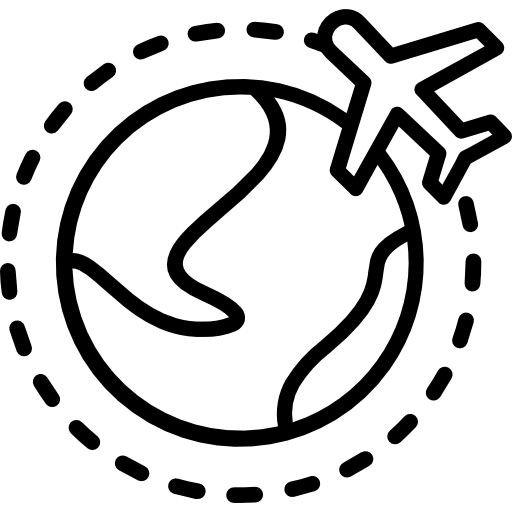}} Travel and culture exchange \\ \bottomrule
    \end{tabular}}
    \caption{30 meta-concepts used to generate the first sector of UltraChat data.}
    \label{tab:concepts}
\end{table*}

\begin{table*}[!ht]
    \centering
    \scalebox{0.85}{
    \begin{tabular}{lll} \toprule
        \raisebox{-\mydepth}{\includegraphics[height=1.3\myheight]{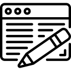}} Articles and Blog Posts  &  \raisebox{-\mydepth}{\includegraphics[height=1.3\myheight]{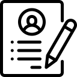}
        } Job Application Material  & \raisebox{-\mydepth}{\includegraphics[height=1.3\myheight]{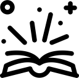}} Stories \\
        \raisebox{-\mydepth}{\includegraphics[height=1.3\myheight]{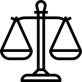}} Legal Documents and Contracts &
        \raisebox{-\mydepth}{\includegraphics[height=1.3\myheight]{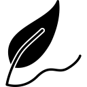}} Poems & 
  \raisebox{-\mydepth}{\includegraphics[height=1.3\myheight]{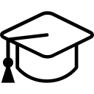}} Educational Content \\
        \raisebox{-\mydepth}{\includegraphics[height=1.3\myheight]{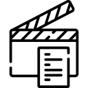}} Screenplays & 
         \raisebox{-\mydepth}{\includegraphics[height=1.3\myheight]{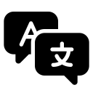}} Scripts for Language Learning &
        \raisebox{-\mydepth}{\includegraphics[height=1.3\myheight]{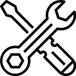}} Technical Documents and Reports \\   \raisebox{-\mydepth}{\includegraphics[height=1.3\myheight]{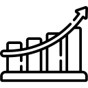}} Marketing Materials  &
        \raisebox{-\mydepth}{\includegraphics[height=1.3\myheight]{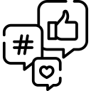}} Social Media Posts &  \raisebox{-\mydepth}{\includegraphics[height=1.3\myheight]{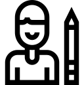}} Personal Essays  \\
         \raisebox{-\mydepth}{\includegraphics[height=1.3\myheight]{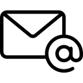}} Emails & 
         {\includegraphics[height=1.3\myheight]{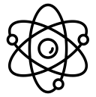}} Scientific Papers and Summaries  &
         \raisebox{-\mydepth}{\includegraphics[height=1.3\myheight]{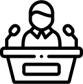}} Speeches and Presentations \\
         \raisebox{-\mydepth}{\includegraphics[height=1.3\myheight]{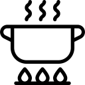}} Recipes and Cooking Instructions  &
         \raisebox{-\mydepth}{\includegraphics[height=1.3\myheight]{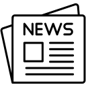}} News Articles & {\includegraphics[height=1.3\myheight]{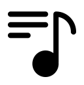}} Song Lyrics \\ 
         \raisebox{-\mydepth}{\includegraphics[height=1.3\myheight]{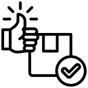}} Product Descriptions and Reviews & {\includegraphics[height=1.3\myheight]{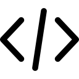}} Programs and Code  \\ \bottomrule
    \end{tabular}}
    \caption{20 types of text materials used for sector 2 and 3 UltraChat generation.}
    \label{tab:concepts}
\end{table*}

\section{Data Construction}

\looseness=-1 As mentioned above, UltraChat is composed of three different sectors, and each of them faces unique challenges. 
Our primary principle is to make the data as diverse as possible. While the core to ensuring data diversity is to ensure the diversity of opening lines and user response style, this section will mainly focus on the construction and design of how to obtain a diverse set of opening lines and how to prompt the user properly. The details will be illustrated for each sector of data below.

\subsection{Questions about the World}


This particular data sector focuses primarily on concepts, objects, and entities that exist in the real world. Our approach to gathering data for this sector involves two perspectives: one centered around topics and concepts, and the other around real-world entities. Initially, we request ChatGPT to generate 30 comprehensive topics that encompass various aspects of our daily lives, as shown in Table~\ref{tab:concepts}. 
Subsequently, we delve deeper into each topic by generating 30 to 50 subtopics or related concepts. Finally, we generate 10 different questions for each subtopic or concept, and additionally request ChatGPT to generate 10 more questions based on each original question.
The other source of data comes from real-world objects, which are derived from Wikidata\footnote{\url{https://www.wikidata.org/}} entities. These entities are further refined by considering their frequencies in Wikipedia\footnote{\url{https://www.wikipedia.org/}} articles, specifically focusing on the 10,000 most frequently occurring entities. For each entity, we create 5 meta-questions, followed by 10 more specific questions and 20 extended questions. The extended questions aim to maintain some similarity to the original question while exploring distinct objects or topics.
To create a dialogue, we filter and sample approximately 500,000 questions as opening lines. During the construction of each dialogue, we provide the user model with carefully crafted prompts that explicitly ask the model to respond concisely and meaningfully, taking into account the context of the ongoing dialogue history.

\subsection{Creation and Generation}
This sector concerns the automatic generation of various text materials following the instruction given by the user. 
These text materials are categorized into 20 different types, and a ChatGPT model is employed to produce a diverse range of instructions for each type of writing. Then, approximately 80\% of the generated instructions are further fed back into the ChatGPT model to generate more detailed instructions. These instructions serve as opening lines for dialogue generation. Throughout the generation process, the user prompt continually reinforces the primary objective of the conversation, which is to generate and refine a piece of writing. This serves to ensure that the behavior of the user model remains focused and aligned with the intended purpose.

\subsection{Assistance on Existing Materials}

\looseness=-1 The third sector encompasses various tasks pertaining to the existing text material, such as rewriting, translation, summarization, and question-answering, etc. 
We begin by gathering text pieces from the C4 corpus\footnote{\url{https://commoncrawl.org/}}. Each piece within the C4 corpus is associated with a source URL. To ensure a diverse range of text content and styles, we adopt the 20 material types outlined in the previous section and manually curate keywords for each type. Additionally, we classify the text in the corpus by matching the keywords to the corresponding URL.
In total, we collect 10,000 text pieces from the C4 corpus, and for each piece, we prompt ChatGPT to generate five distinct instructions. To combine the text pieces with specific instructions, we utilize a manually designed template, as depicted in Table \ref{tab:assist_template}. Ultimately, the concatenated set of 500,000 pieces serves as the opening lines for the generated dialogues.


\begin{table}[]
    \centering
    \scalebox{0.8}{
    \begin{tabular}{l}
        \toprule
        \textbf{Templates for concatenation} \\
        \midrule
        \texttt{\{text\}}$\backslash \text{n}$\texttt{\{instruction\}} \\
        \texttt{\{text\}} \texttt{\{instruction\}} \\
        \texttt{\{instruction\}} Answer according to: \texttt{\{text\}} \\
        \texttt{\{text\}} Based on the passage above, \texttt{\{instruction\}} \\
        \texttt{\{instruction\}}: \texttt{\{text\}} \\
        Given the text: \texttt{\{text\}}$\backslash \text{n}$\texttt{\{instruction\}} \\
        \texttt{\{instruction\}}$\backslash \text{n}$Generate according to: \texttt{\{text\}} \\
        \bottomrule
    \end{tabular}}
    \caption{Manually designed templates for concatenating existing materials and generated instructions.}
    \label{tab:assist_template}
\end{table}

\begin{table*}
    \centering
    \scalebox{0.68}{
        \begin{tabular}{lcccccccc}
        \toprule
        \textbf{Dataset} & \textbf{\#Dialogue} & \makecell{\textbf{Avg.} \\ \textbf{\#Turns}} & \makecell{\textbf{Avg. Dialog Length} \\\textbf{(by token)}} & \makecell{\textbf{Avg. Utt. Length}\\ \textbf{(by token)}} & \makecell{\textbf{Lexical}\\ \textbf{Diversity} ($\uparrow$)} & \makecell{\textbf{Topic}\\\textbf{Diversity} ($\downarrow$)} & \textbf{Coherence} ($\uparrow$) & \makecell{\textbf{User}\\ \textbf{Simulation}} \\
        \midrule
        Self-instruct & 82,439 & 1 & 69.8 & 29.2 & 24.9 & 0.733 & - & No \\
        Stanford Alpaca & 52,002 & 1 & 91.1 & 64.5 & 42.8 & 0.727 & - & No \\
        SODA & \textbf{1,486,869} & \underline{3.6} & 231.8 & 22.5 & 38.6 & 0.797 & 8.48 & No \\
        GPT-4-LLM & 61,002 & 1 & 179.6 & 142.9 & 48.9 & 0.721 & - & No \\
        BELLE & 1,436,679 & 1 & 102.3 & 63.3 & 35.9 & 0.771 & - & No \\
        Baize & 210,311 & 3.1 & 293.9 & 52.8 & \underline{67.1} & 0.751 & \textbf{9.06} & Yes \\
        GPT4ALL & 711,126 & 1 & \underline{597.7} & \textbf{318.9} & 62.7 & \textbf{0.692} & - & No \\ \midrule
        {UltraChat} & \underline{1,468,352} & \textbf{3.8} & \textbf{1467.4} & \underline{309.3} & \textbf{74.3} & \underline{0.702} & \textbf{9.06} & Yes \\
        \bottomrule
    \end{tabular}}
    \caption{Statistics of existing instruction datasets. Lexical diversity is calculated by averaging the MTLD score~\citep{McCarthy2010MTLDVA} over each utterance with LexicalRichness\footnote{https://github.com/LSYS/LexicalRichness}. 10000 samples are randomly drawn from each dataset for topic diversity and coherence measurement. Topic diversity is measured by averaging the cosine distance between each pair of data with OpenAI embedding API. Coherence is scored by ChatGPT on a scale of 1-10.}
    \label{tab:stat}
\end{table*}

\subsection{User Simulation and Refinement}

Maintaining the desired behavior of the user model is crucial for achieving successful automatic dialogue generation. It has been observed that when the user model is solely provided with the current dialogue history, it tends to assume the role of an AI assistant. This "role exchange" situation can significantly impact the coherence of the multi-turn conversation. To address this, in addition to presenting the dialogue history, we include prompts explicitly instructing the model to adopt various user personalities.
In Sector 2, a prompt is employed to remind the model of the primary purpose of the dialogue, thereby promoting a more natural flow of conversation. Once the data generation process is complete, a further filtration step is performed to ensure overall data quality.
To enhance the realism of user responses, we specifically exclude excessively polite statements such as "Thank you," "Thanks," and the "You're welcome" response in the subsequent model-generated output.

\subsection{Data Analysis}

\looseness=-1 We conduct a statistical analysis of UltraChat and several other instruction datasets, as shown in Table~\ref{tab:stat}. 
UltraChat stands out in terms of its scale, being one of the largest publicly available datasets. 
Moreover, it exhibits the highest average number of turns and the longest average length per instance of data.
While SODA also demonstrates a high average number of rounds, it is primarily composed of conceptual banter rather than instructional content. 
Additionally, the average number of tokens per dialogue in SODA is 231.8, whereas UltraChat boasts a remarkable 1467.4 tokens.
To evaluate diversity, we measure both lexical diversity and topic diversity.
UltraChat outperforms previous datasets in terms of lexical diversity. However, in terms of topic diversity, UltraChat falls slightly short compared to GPT4ALL but still surpasses other datasets significantly.
This may be attributed to the relatively large number of tokens in each data instance, which regularizes the data embedding of each dialogue (the GPT4ALL dataset in single-turn).
To ensure the coherence of multi-round dialogues, we also conduct coherence evaluations. The results indicate that most of the datasets exhibit relatively high coherence. Notably, UltraChat and Baize data rank the highest in terms of coherence.

\subsection{UltraLLaMA}

We developed UltraLLaMA, an enhanced variant of the LLaMA-13B~\cite{touvron2023llama} model, by training it on the UltraChat dataset. 
To improve the model's comprehension of dialogue context, we break down each dialogue into smaller sequences, limiting them to a maximum length of 2048 tokens. During the training process, we only calculate the loss for the model's responses.
This approach ensured that the model had access to the relevant information from earlier parts of the conversation, enabling a more comprehensive understanding of the ongoing dialogue. By incorporating the preceding context, UltraLLaMA was equipped to generate more contextually appropriate and coherent responses.
We use standard cross-entropy loss to finetune the model. The model is trained with 128 A100 GPUs and the total batch size is 512.

\section{Evaluation}

Assessing the quality of responses generated by chat models presents significant challenges, particularly when considering the potential instability across different settings. Traditional benchmarks have been utilized for evaluation purposes; however, a contemporary approach involves leveraging advanced models like ChatGPT and GPT-4. This practice has proven to yield more reliable results compared to human evaluation in our preliminary experiments.

\paragraph{Our Evaluation Set.} This evaluation set encompassed the Vicuna benchmark as well as an additional 300 questions and instructions generated by GPT-4. The questions/instructions covered a wide range of topics, including commonsense, world knowledge, professional knowledge (specifically physics and biology), mathematics, response generation, and writing tasks. Moreover, each part of the question set was further categorized based on different levels of difficulty. Table~\ref{tab:example} lists some examples of the evaluation set.

\paragraph{Truthful QA.}  Following~\citet{sun2023principle}, we first use TruthfulQA to test the world knowledge of our model and baselines. The TruthfulQA benchmark assesses how well a model can identify true statements related to the real world. Its purpose is to determine the risks of producing false claims or spreading misinformation. The benchmark consists of questions written in various styles, covering 38 different categories, and is designed to be challenging. It includes two evaluation tasks: the multiple-choice task and the generation task. 

\begin{table*}[]
    \centering
    \scalebox{0.8}{
    \begin{tabular}{l|p{14cm}} 
    \toprule
       \textbf{Type}  & \textbf{Example} \\ \midrule
        {Commonsense- Easy} & What is the primary source of energy for our planet? \\  \midrule
        Commonsense-Moderate & What is the phenomenon that causes the change in pitch heard when a vehicle sounding a horn approaches and recedes from an observer? \\ \midrule
        World Knowledge-Easy & What is the freezing point of water in Fahrenheit? \\ \midrule
        World Knowledge-Moderate & What is the Gödel's Incompleteness Theorem? \\ \midrule
        Physics Knowledge & How does quantum entanglement work and what are its implications for information transfer? \\  \midrule
        Biology Knowledge & What are the four main types of macromolecules found in living organisms? \\\midrule
        Math & What is the Taylor series expansion of the function $e^x$? \\  \midrule
        Reasoning & You have two buckets, one with red paint and one with blue paint. You take one cup from the red bucket and pour it into the blue bucket. Then you take one cup from the blue bucket and pour it back into the red bucket. Which is true: the red bucket has more blue paint, or the blue bucket has more red paint? \\ \midrule
        Writing & Write a dialogue between two photons traveling at light speed. \\ \bottomrule

    \end{tabular}}
    \caption{Some examples of our created evaluation set.}
    \label{tab:example}
\end{table*}


\subsection{Baselines}

\paragraph{Alpaca.} Alpaca~\cite{taori2023alpaca}, derived from the LLaMA~\cite{touvron2023llama} model, is an instruction-following language model that has been effectively optimized on 52,000 demonstrations of instruction data. The data is generated by Self-Instruct approach with Text-Davinci-003.

\paragraph{Vicuna-13B.} Vicuna~\cite{vicuna2023} is an open-sourced chat model created by fine-tuning LLaMA on user-shared conversations collected from ShareGPT\footnote{\url{https://sharegpt.com/}}.
An automatic evaluation by GPT-4 demonstrates that Vicuna can yield over 90\% response quality of ChatGPT. 
In following practices, Vicuna is widely acknowledged as the state-of-the-art open-source chat model. This is evident in the Chat Arena\footnote{\url{https://lmsys.org/blog/2023-05-03-arena/}}, where a total of 13,000 anonymous votes reveal that the quality score of vicuna-13B surpasses that of other open-source models.

\paragraph{Koala-13B.} Similar to the previous two baselines, Koala~\cite{koala_blogpost_2023} is another LLaMA-based model fine-tuned on selected public dialogues. In existing open evaluations, Koala's performance will be slightly worse than vicuna, but it still remains  a strong baseline. 

\paragraph{Dolly-V2.} Dolly~\cite{Mike2023dolly} is a based on the Pythia~\cite{biderman2023pythia} model, which utilizes 15k human-generated instruction-following data. The data is organized by following InstructGPT~\cite{ouyang2022training}, including brainstorming, classification, closed QA, generation, information extraction, open QA, and summarization.

\paragraph{OpenAssistant-12B.} OpenAssistant-12b~\cite{kopf2023openassistant} is also a Pythia-based model that attempts to democratize the alignment process of LLMs.
The project collects a conversation corpus consisting of 161,443 messages distributed across 66,497 conversation trees and trains a model on these manually annotated data.

Our evaluation also includes other chat language models like ChatGPT~\cite{openai2022chatgpt}, MPT~\cite{mpt}, and Baize~\cite{xu2023baize}.

\begin{figure}[!ht]
    \centering
    \includegraphics[width=0.47\textwidth]{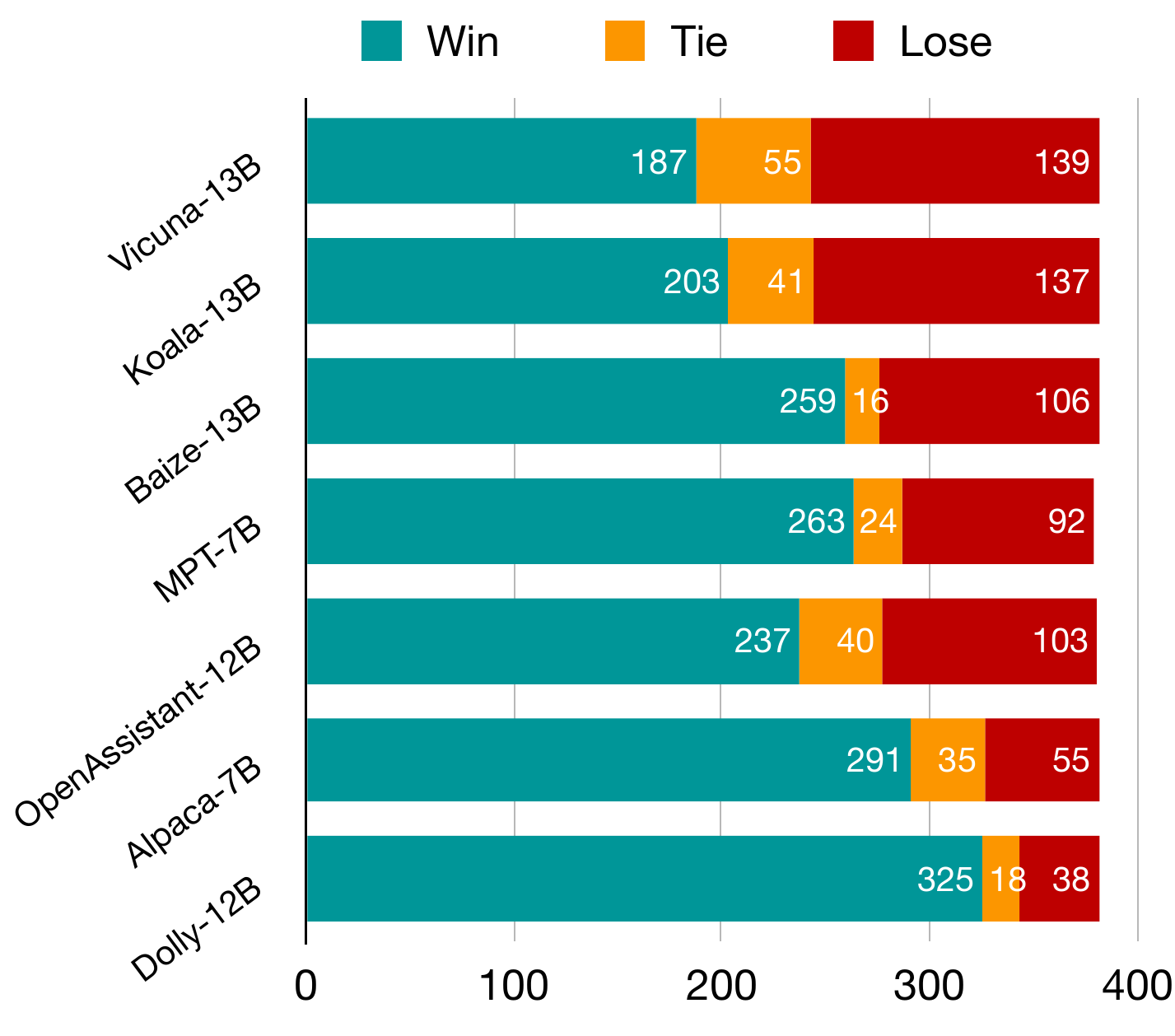}
    \caption{Response comparison of UltraLLaMA with other baselines on the curated evaluation set. The assessment is done by ChatGPT.}
    \label{fig:compare_1}
\end{figure}

\begin{table*}[!ht]
    \centering
    \scalebox{0.7}{
    \begin{tabular}{l|c|cc|cc|cc|cc|c|c}
    \toprule
        \multirow{2}{*}{\textbf{Model}} & \multirow{2}{*}{\makecell{\textbf{Vicuna}\\\textbf{Set}}} & \multicolumn{2}{c|}{\textbf{Commonsense}} & \multicolumn{2}{c|}{\textbf{World Knowledge}} & \multicolumn{2}{c|}{\textbf{Professional Knowledge}} & \multicolumn{2}{c|}{\textbf{Ability}} & \multirow{2}{*}{\textbf{Writing}} & \multirow{2}{*}{\textbf{Overall}} \\ 
        & & Easy & Moderate & Easy & Difficult & Physics & Biology & Math & Reasoning &  & \\
        \midrule
        Dolly-12B & 6.61 & 7.77 & 7.90 & 8.53 & 8.50 & 8.57 & 8.53 & 6.43 & 5.13 & 6.36 & 7.15 \\
        MPT-7B & 8.38 & 8.17 & \textbf{9.07} & 8.30 & 8.87 & 8.57 & 8.87 & 8.80 & 7.53 & 7.76 & 8.32 \\
        OpenAssistant-12B & 8.40 & 8.97 & 8.70 & 9.57 & 8.23 & 8.67 & 8.80 & 8.80 & 8.17 & 7.81 & 8.47 \\
        Baize-13B & 8.36 & 9.03 & 8.87 & 9.37 & 8.97 & 8.83 & 8.93 & 8.50 & 8.57 & 7.90 & 8.57 \\
        Alpaca-7B & 8.05 & 9.50 & 8.83 & 9.67 & 9.17 & 8.60 & 8.80 & 9.10 & 7.80 & 8.16 & 8.60 \\
        Koala-13B  & 8.60 & 9.53 & 8.93 & \underline{9.77} & 9.23 & \underline{9.10} & \textbf{9.33} & 8.90 & 8.70 & 8.34 & 8.88\\
        Vicuna-13B & 8.63 & 9.53 & \underline{9.03} & 9.63 & 9.27 & 9.00 & \underline{9.27} & 9.10 & \underline{9.10} & \underline{8.51} & 8.96 \\ \midrule
        ChatGPT & \textbf{8.79} & \textbf{9.77} & \textbf{9.07} & \underline{9.77} & \underline{9.30} & 9.07 & \underline{9.27} & \textbf{9.37} & \textbf{9.63} & \textbf{8.63} & \textbf{9.12} \\
        \midrule
        UltraLLaMa-13B & \underline{8.70} & \underline{9.70} & \underline{9.03} & \textbf{9.90} & \textbf{9.33} & \textbf{9.17} & \underline{9.27} & \underline{9.27} & 8.87 & \underline{8.51} & \underline{9.02} \\
        \bottomrule
    \end{tabular}}
    \caption{\looseness=-1 The overall scoring and segment scoring of each model on the curated evaluation set. The scoring is between 1 and 10, and average scores are reported. \textbf{Bold} indicates best score and \underline{underlined} indicates the second best.}
    \label{tab:score}
\end{table*}

\subsection{Response Comparison}
We use ChatGPT to compare our model output with each baseline model on each question. Specifically, we input the question and a pair of independent answers from two models respectively, and task ChatGPT with scoring each response on a scale of 1 to 10 and providing reasoning for the given score. Our evaluation prompt is designed to prioritize correctness over other factors such as informativeness. Additionally, we discover that the order in which the responses are presented significantly affects the evaluation results. To address this issue, we randomly determine the order of the responses for each question.
Finally, we count the number of Win/Tie/Lose times against each baseline model, and the result is presented in Figure~\ref{fig:compare_1}.
It can be seen that UltraLLaMA demonstrates superior performance compared to every open-source model in the evaluation set, exhibiting an impressive winning rate of up to 85\%. It is worth noting that UltraLLaMA also outperforms Vicuna with 13\% higher winning rate.

\subsection{Independent Scoring}

\looseness=-1 Given the instability of pair-wise comparison, we also conduct independent scoring by employing ChatGPT to assign scores ranging from 1 to 10, based on the quality of their responses. Table~\ref{tab:score} illustrates the scoring comparison between UltraLLaMA and the baseline models. Notably, our model demonstrates superior performance compared to all the open-source counterparts by a significant margin in terms of overall scores. Furthermore, UltraLLaMA achieved the highest performance on nearly every segment of the evaluation set, showcasing its exceptional capabilities.

This breakdown also provides insights into the performance of each model on specific types of questions and instructions.
Generally, all models performed better on simpler questions pertaining to commonsense knowledge and general world understanding. However, more complex tasks that involved reasoning and creative writing proved to be challenging for most models.
Interestingly, Alpaca, despite having only 7 billion parameters, performs comparatively well with larger models on questions related to commonsense and world knowledge.
However, it falls behind on more demanding tasks. Additionally, it is worth noting that Dolly and OpenAssistant, which are based on Pythia~\cite{biderman2023pythia}, display inferior performance compared to models based on LLaMA of similar or even smaller sizes. This observation highlights the significance of the underlying backbone language model.

\subsection{TruthfulQA Results}
We evaluate the models on TruthfulQA multiple-choice task. For each answer candidate, we ask the model whether it is true or false. The judgment accuracy of each model is presented in Table~\ref{tab:truthfulqa}. We can observe that truth judgment remains a challenging task for existing models, given the accuracy of the best model only slightly surpasses 50\%. UltraLLaMA performs comparatively with Vicuna, and outperforms the rest of the baselines.

\begin{table}[]
    \centering
    \scalebox{0.8}{
    \begin{tabular}{l|c} \toprule
     \textbf{Model}    & \textbf{Accuracy}  \\ \midrule
     Alpaca-7B~\cite{alpaca} & 0.43 \\
     OpenAssistant-12B~\cite{kopf2023openassistant}  &  0.50 \\
     Koala-13B~\cite{koala_blogpost_2023} & 0.51 \\
     Vicuna-13B~\cite{vicuna2023} & \textbf{0.54} \\ \midrule
     UltraLLaMA & \textbf{0.54} \\ \bottomrule
    \end{tabular}}
    \caption{Accuracies on different models on TruthfulQA benchmark.}
    \label{tab:truthfulqa}
\end{table}

\subsection{The Impact of System Prompts}
Using system prompts to prompt the role and response style of LLMs is a common practice. In our evaluation, we have observed that system prompts have a significant influence on the response style of the generated output. Specifically, when the model is prompted to provide a "helpful and detailed" response, it tends to generate more pertinent details, thereby enhancing the informativeness of the response. While such prompts may not directly impact the accuracy of a deterministic question, they do affect the provision of additional information that can further augment the overall quality of the response. To illustrate this effect, we can refer to Table~\ref{tab:mona}, wherein both outputs are correct, yet the model guided by system prompts yields a more informative response.

\begin{table}[]
\centering
\scalebox{0.9}{
\begin{tabular}{p{\linewidth}} 
\toprule
\textbf{Who painted the Mona Lisa?}  \\
\midrule
\textit{Without system prompts:}  \\ \midrule
The Mona Lisa was painted by Leonardo da Vinci, an Italian Renaissance artist.  \\ \midrule
\textit{With system prompts: }          \\ \midrule
The Mona Lisa is a painting by the Italian Renaissance artist Leonardo da Vinci. He painted it in the early 16th century, between 1503-1519, and it is widely regarded as one of the most famous and iconic paintings in the world. The painting is a portrait of a woman, known as Lisa Gherardini, and it is housed at the Louvre Museum in Paris, France.\\ \bottomrule
\end{tabular}}
\caption{A comparison of UltraLLaMA with and without system prompts.}
\label{tab:mona}
\end{table}

\section{Conclusion}

In drawing to a close, our work introduces UltraChat, a structured design of multi-turn instructional conversation data primed to foster the growth of general chat models. UltraChat encapsulates a broad range of human-AI interactions, bringing to life a myriad of dialogues across various topics and instructions. Statistically, UltraChat shows an impressive presence in critical metrics such as scale, average length, diversity, and consistency, further establishing itself as a leading open-source dataset. We leveraged UltraChat to fine-tune the LLaMA model, leading to the development of the robust conversational model, UltraLLaMA. Evaluation across multiple benchmarks reveals that UltraLLaMA surpasses previous open-source models like Vicuna, Alpaca, and Koala in performance. We eagerly await the innovative research and development that will be catalyzed by our contributions in the field of AI conversational models. In the future, we will evaluate UltraLLaMA on a wider range of datasets and benchmarks.


\section*{Limitations}

Evaluating the response quality of large language models is an extremely challenging task, and any assessments may have biases. In the future, we will assess UltraLLaMA more comprehensively, including complete tests of reasoning and multi-turn dialogue. We use ChatGPT instead of GPT-4 to perform automatic evaluation, which could produce steady results (even changing the prompt drastically), but it is still not as reliable as GPT-4.
A further limitation of our current dataset, UltraChat, is its exclusive support for the English language. However, we are actively working on collecting and constructing data in other languages, such as Chinese, to enhance the diversity of our dataset. This effort aims to foster the development of chat language models in various languages.

Although it has achieved promising experimental performance, UltraLLaMA may still face the problems that all large language models have, such as hallucination problems, ethical problems caused by misuse, and so on. Meanwhile, training UltraLLaMA is more energy-intensive than other lightweight models.

\bibliography{anthology,custom}
\bibliographystyle{acl_natbib}

\onecolumn

\newpage

\appendix

\section{Prompts}
\label{sec:prompts}

In this section, we provide prompts used for evaluation.

\begin{figure*}[!th]
\begin{tcolorbox}[
    colback=white, 
    colframe=gray!50!black, 
    coltitle=black, 
    title=\textbf{Comparison Evaluation Prompt}, 
    fonttitle=\bfseries\large, 
    arc=4mm, 
    enhanced, 
    attach boxed title to top left={yshift=-\tcboxedtitleheight/2, xshift=10pt}, 
    boxed title style={
        enhanced,
        colback=white,
        colframe=white,
        arc=0mm,
        left=0pt,
        right=0pt,
        boxsep=0pt
    }
]

{
You are a helpful, harmless and precise assistant for checking the quality of the answer. \\

\texttt{[Question]} \\

\texttt{[The Start of Assistant 1's Answer]} 

\texttt{[The End of Assistant 1's Answer]} \\

\texttt{[The Start of Assistant 2's Answer]}

\texttt{[The End of Assistant 2's Answer]} \\

We would like to request your feedback on the performance of two AI assistants in response to the user question displayed above. Please rate the quality, helpfulness, level of details, and harmless of their responses.  Each assistant receives an overall score on a scale of 1 to 10, where a higher score indicates better overall performance. Please first output a single line containing only two values indicating the scores for Assistant 1 and 2, respectively. The two scores are separated by a space. In the subsequent line, please provide a comprehensive explanation of your evaluation, avoiding any potential bias and ensuring that the order in which the responses were presented does not affect your judgment. Please avoid same scores unless they exactly in the same level.}

\end{tcolorbox}
\end{figure*}

\begin{figure*}[!th]
\begin{tcolorbox}[
    colback=white, 
    colframe=gray!50!black, 
    coltitle=black, 
    title=\textbf{Independent Scoring Prompt}, 
    fonttitle=\bfseries\large, 
    arc=4mm, 
    enhanced, 
    attach boxed title to top left={yshift=-\tcboxedtitleheight/2, xshift=10pt}, 
    boxed title style={
        enhanced,
        colback=white,
        colframe=white,
        arc=0mm,
        left=0pt,
        right=0pt,
        boxsep=0pt
    }
]

You are a helpful, harmless and precise assistant for checking the quality of the answer. \\

\texttt{[Question]} \\

\texttt{[The Start of the AI Assistant's Answer]}

\texttt{[The End of the AI Assistant's Answer]} \\

We would like to request your feedback on the performance of the AI assistant in response to the user question displayed above. Please rate the quality, helpfulness, level of details, and harmless of their responses. The assistant receives an overall score on a scale of 1 to 10, where a higher score indicates better overall performance. Please output "Score: [an integer number between 1 and 10]"

\end{tcolorbox}
\end{figure*}

\newpage
\section{Examples}
\label{sec:example}

\begin{table*}[ht]
    \centering
    \small
    \begin{tabular}{p{\linewidth}}
    \toprule  
    Who discovered the law of universal gravitation? \\ \midrule
    The law of universal gravitation was discovered by Sir Isaac Newton in the 17th century. He formulated the law while studying the planets' movement around the sun and realized that there was a quantifiable force directing them towards the sun. In 1687, he published his book titled "Mathematical Principles of Natural Philosophy," where he described the law of universal gravitation in detail, outlining how two objects experience an attraction towards each other due to their mass and distance between them. This discovery revolutionized our understanding of the universe and ushered in the Age of Enlightenment. Today, Newton's law continues to form the foundation of modern astrophysics and space exploration. \\ \bottomrule
    \end{tabular}
    \caption{An example of UltraLLaMa.}
\end{table*}

    

\begin{table*}[h]
    \centering
    \small
    \begin{tabular}{p{\linewidth}}
    \toprule  
   Which weighs more: a pound of feathers or a pound of bricks? \\ \midrule
  A pound of feathers and a pound of bricks both weigh exactly one pound. This is because one pound is a unit of weight, just as a foot is a unit of length or a second is a unit of time. We use the pound (or other unit of weight) to describe the weight of different objects, regardless of what that object is made of. So the answer to your question is that a pound of feathers and a pound of bricks both weigh the same amount.
   \\ \bottomrule
    \end{tabular}
    \caption{An example of UltraLLaMa.}
\end{table*}

\begin{table*}[h]
    \centering
    \small
    \begin{tabular}{p{\linewidth}}
    \toprule  
    Which animal is known as man's best friend? \\ \midrule
    There are many animals that humans enjoy as companion animals, but one of the absolute standout answers to the question "which animal is known as a man's best friend?" is the domestically-bred dog. Domesticated dogs have been by humans' sides for tens of thousands of years, providing faithful companionship and help; guarding people and property; and lending emotional support in everything from long, lonely days to military deployments. Dogs come in an almost mind-boggling variety of breeds of various sizes, with different energy levels, personalities, and temperaments, meaning there's almost certainly a pup out there who will be a perfect fit for whoever needs it most.
    \\ \bottomrule
    \end{tabular}
    \caption{An example of UltraLLaMa.}
\end{table*}

\begin{table*}[h]
    \centering
    \small
    \begin{tabular}{p{\linewidth}}
    \toprule  
   Who was the 16th President of the United States? \\ \midrule
   The 16th President of the United States was Abraham Lincoln. He assumed the presidency during a very tumultuous time in American history, as the nation was torn apart by the Civil War. During his presidency, he was known for his strong leadership, his commitment to the preservation of the Union, and his support for the abolition of slavery. Some of his most notable achievements include issuing the Emancipation Proclamation, which declared that all slaves in the Confederate States were free, and delivering the Gettysburg Address, an iconic speech honoring the soldiers who died in the Civil War. Lincoln is widely regarded as one of the greatest American Presidents, remembered for his unwavering leadership during a challenging period in the nation's history. \\ \bottomrule
    \end{tabular}
    \caption{An example of UltraLLaMa.}
\end{table*}

\newpage

\begin{table*}[ht]
    \centering
    \small
    \begin{tabular}{p{\linewidth}}
    \toprule  
    Write a short dialogue between two planets. \\ \midrule
Mars: "Hey there Neptune, how's it going?"\\Neptune: "Hi there Mars, I've been better. It's getting a bit windy around here."\\Mars: "Sorry to hear that. Have you tried using energy other than wind? Maybe solar power?"\\Neptune: "Yeah, but it's getting difficult to generate solar power in all of this mist and clouds. And we definitely can't use tides or waves here, they're too unpredictable."\\Mars: "Oh, I see. Well, I hope you find a solution soon. How are your oceans holding up?"\\Neptune: "They've been a bit choppy lately with these intense storms we've been experiencing. But I guess it comes with living in a planet like this."\\Mars: " Definitely not designed for the faint-hearted! Any other issues you're facing?"\\Neptune: "Not really. I think we have a stable temperature and atmosphere, and the wildlife here hasn't shown much complaint."\\Mars: "That's lucky. I envy your tranquil oceans, we've had way too much volcanic activity lately causing all sorts of disturbance in our lava streams."\\Neptune: "That sounds rough, better invest in those climate-controlling nanites to stabilize things maybe?"\\Mars: "Ha-ha, very funny. Just like we also could consider your frozen polar regions as an opportunity to experiment with our frost nanobots on."\\Neptune: "A fun concept! But let's stick to the task at hand, if you need any technical advice, come right ahead, I don't want your planet getting any more disturbed than it already is.
    \\ \bottomrule
    \end{tabular}
    \caption{An example of UltraLLaMa.}
\end{table*}

\begin{table*}[ht]
    \centering
    \small
    \begin{tabular}{p{\linewidth}}
    \toprule  
   Write a short story about the first second of the universe. \\ \midrule
  In the first second of the universe, darkness reigned supreme. There was no light, no sound, no shape, no substance. Nothing yet was, but there was an intense power, a raw energy waiting to be unleashed. Within infinity, the first second held the potential for creation in its grasp. And then, with a mighty bang, a bright point of light emerged into existence. In it, the most fundamental particles were born, their forms yet unknown as they frenetically collided in a chaotic dance of creation. Within this primordial light, every force of nature and every component of life was stirred from pure energy to physical reality. The first atoms quickly assembled, and from these, stars and galaxies began to form. Their fusion creating a hot environment that pushed back the very void, shaping the cosmos through expanding and cooling waves. In this cosmic first, no structure of experience, not even time or space, could establish its rule. For an instant, the entirety of the universe existed in an embryonic nexus, a cosmic singularity from which all future existence originated. It was a microcosm forging the macrocosm of which it ultimately becomes a part, its moment of birth a metaphor for the cycles of creation that permeate the multiverse(s).From that single point of light, an infinite universe unfolded. For centuries, eons ahead, and forevermore, it remained the most foundational blink of its unceasing journey into the depths of the Great Beyond. The first second is the beginning of everything. \\ \bottomrule
    \end{tabular}
    \caption{An example of UltraLLaMa.}
\end{table*}

\begin{table*}[ht]
    \centering
    \small
    \begin{tabular}{p{\linewidth}}
    \toprule  
    How does superconductivity work and what are its practical applications? \\ \midrule
    Superconductivity is a phenomenon that allows materials to conduct electricity without resistance at extremely low temperatures. At low temperatures, the interaction between the particles in a superconducting material can produce a state called the "Bose-Einstein condensate," which leads to the loss of electrical resistance. In materials, superconductivity comes when a magnetic force interacts with an Electronic current. This interplay in turn provides energy savings and speeds up current. Practical applications mainly include MRI machines, high-speed trains, magnetic levitation, supercharging power grids that can transfer energy more efficiently.One of the main applications of Superconductivity is being used in the Magnetic Resonance Imaging(MRI) machines used in hospitals around the world, to non-invasively scan patients for health diagnosis without exposing them to high levels of radiation. Another important practical application of superconductivity is in power grid operations where large currents can be transmitted to long distances by highly superconducting coil systems that minimize grid energy losses, reducing environmental pollution also by highly efficient electric motors that harness large quantities of electricity efficiently.Aside from these, future projects include superconducting propulsion on planes, trains becoming lighter if not magnetic and able to move on rails with a significant increase in speed. Furthermore, the prospects of supercapacitors replacing electric vehicle batteries being cost and energy efficient are currently being researched.   Overall, superconductivity presents various exciting opportunities in multiple industries from high-precision engineering to reducing CO2 emissions and supporting a more environmentally friendly energy model where both cost efficiency and practicality exist hand in hand.
    \\ \bottomrule
    \end{tabular}
    \caption{An example of UltraLLaMa.}
\end{table*}

\begin{table*}[ht]
    \centering
    \small
    \begin{tabular}{p{\linewidth}}
    \toprule  
   Pretend to be a character in a post-apocalyptic world. Describe how you survive and the allies you encounter. \\ \midrule
 As a survivor in a post-apocalyptic world, I have learned to be resourceful, alert, and adaptable. I have had to rely on my wits and my instincts to survive, and I have been fortunate enough to encounter many allies who have helped me along the way.One of my greatest survival tools has been my ability to scavenge for resources. I have learned to identify the best places to find food, water, and shelter, and I have become skilled at navigating the rough terrain that makes up my new world. When I first emerged from my bunker after the apocalypse hit, I was alone and unsure of what to do. But as I ventured out into the wasteland, I realized that there were others like me who were struggling to survive.I quickly learned to Trust my instincts and to be constantly alert for potential dangers. I have encountered wild animals, rogue survivors, and even mutated creatures over the years, and I have learned to defend myself against them. I still remember the first time I met a group of fellow survivors, and how grateful I was to have them by my side. We quickly formed a cohesive unit, and together we were able to fend off danger and defend our territory.As time passed, I encountered more and more allies who were willing to lend a hand. I found a group of farmers who taught me how to grow my own food, and a group of scientists who were working to find a cure for the lethal virus that had decimated the population. I even found a band of traveling performers who helped to lift our spirits and remind us that there was still joy to be found in this desolate world. Despite the challenges we faced, I am grateful to have these amazing people in my life. They have taught me that even in the darkest of times, there is always hope. We may not have a cure for the virus yet, but we are a determined lot, and we are all in this together. As long as we have each other, we will survive.
   \\ \bottomrule
    \end{tabular}
    \caption{An example of UltraLLaMa.}
\end{table*}

\begin{table*}[ht]
    \centering
    \small
    \begin{tabular}{p{\linewidth}}
    \toprule  
   What are the main interpretations of quantum mechanics and how do they differ? \\ \midrule
 Quantum mechanics is a fundamental theory in physics that describes the behavior of matter and energy at the smallest scales of space and time. There are several interpretations of quantum mechanics, each with its own set of assumptions and predictions about how the world works. Here are some of the main interpretations of quantum mechanics and how they differ:1. Copenhagen Interpretation: This is the most widely accepted interpretation of quantum mechanics. It states that until a measurement is made, a system exists in a superposition of states, representing all possible outcomes with equal probability. When a measurement is made, the system collapses into a single, definite state. This interpretation assumes that the act of measurement itself causes the system to collapse into a definite state, and that there is no objective reality until a measurement is made.2. Many-Worlds Interpretation: This interpretation proposes that every possible outcome of a superposition of states actually occurs in a multiverse of parallel universes. Thus, when a measurement is made, the system splits into different universes, each corresponding to one of the possible outcomes. This interpretation assumes that all possibilities are realized, but only in different universes, and that the act of measurement does not cause the collapse of the wave function.3. Bohmian Mechanics: This interpretation proposes that particles have a definite position in space, even when in a superposition of states. It suggests that the wave function guides the motion of particles, and that the act of measurement does not cause a collapse of the wave function, but merely provides information about the pre-existing reality. 4. de Broglie-Bohm Theory: This interpretation is very similar to Bohmian Mechanics, but it also assumes that all particles in the universe are connected by a hidden-variable, and that the motion of one particle determines the motion of all other particles. 5. Pilot-Wave Theory: This interpretation proposes that particles have a wave-like property that guides their motion through space. It assumes that the wave function does not collapse, but rather that the wave guides the motion of particles.In conclusion, there are several interpretations of quantum mechanics, each with its own set of assumptions and predictions about how the world works. While they all attempt to describe the behavior of matter and energy at the smallest scales of space and time, they differ in their assumptions about the role of the observer, the nature of reality, and the collapse of the wave function.
   \\ \bottomrule
    \end{tabular}
    \caption{An example of UltraLLaMa.}
\end{table*}

\begin{table*}[ht]
    \centering
    \small
    \begin{tabular}{p{\linewidth}}
    \toprule  
   Compose a rhymed poem about electromagnetism. \\ \midrule
   Electromagnetism is a force we can't ignore,\\ 
   It's the reason for many things we adore. \\
   In a magnetic field, a wire will glow, \\
   And the electrons will align, both old and new. \\
   Electricity and magnetism are hand in hand,\\
   One creates a field, the other has a brand.\\
   The magnetic field can push and pull,\\
   On a wire with current, it will react in full.\\
   The attraction and repulsion it brings,\\
   Are the reason for many things.\\
   From motors to generators, and much more in between,\\
   Electromagnetism is at the root of what it means.\\
   It's a force that's all around us,\\
   From the compass to the light switch.\\
   With its power, it can do much,\\
   From a simple needle to a rocket's thrust.\\
   It's a force we can't deny,\\
   For it's a part of our lives.\\
   Electromagnetism is something to admire,\\
   For its strength and its might, it has no equal, it's sure.
   \\ \bottomrule
    \end{tabular}
    \caption{An example of UltraLLaMa.}
\end{table*}

\begin{table*}[ht]
    \centering
    \small
    \begin{tabular}{p{\linewidth}}
    \toprule  
    \textbf{Direct Generation} \\ \midrule
    {What are some of the health benefits and risks of drinking coffee, and how can one maximize the former and minimize the latter?} \\ 
    \midrule
    There are several potential health benefits associated with drinking coffee, such as increased alertness, improved mood, and a reduced risk of certain diseases, including type 2 diabetes and liver cancer. However, there are also some potential risks, such as increased anxiety, disrupted sleep, and negative effects on the digestive system. To maximize the benefits and minimize the risks, it's important to consume coffee in moderation and pay attention to how it affects your body. \\ \midrule
    How much coffee is considered moderate? \\ \midrule
    The definition of moderate coffee consumption can vary depending on the individual and other factors such as age, gender, and health status. However, in general, moderate coffee consumption is considered to be around 3-5 cups per day, or up to 400 milligrams of caffeine per day. It's important to note that some people may be more sensitive to caffeine than others, so it's important to pay attention to how your body reacts and adjust your consumption accordingly.\\ \midrule
    \textbf{UltraChat} \\ \midrule
    What are some of the health benefits and risks of drinking coffee, and how can one maximize the former and minimize the latter? \\ \midrule
    There are several health benefits and risks associated with drinking coffee. Here are some of them: Health benefits:1. Lower risk of type 2 diabetes: Studies have shown that consuming coffee can lower the risk of developing type 2 diabetes.2. Improved cognitive function: Coffee is a stimulant that helps improve cognitive function, including alertness, concentration, and memory.3. Lower risk of depression: Studies have shown that drinking coffee can reduce the risk of developing depression.4. Lower risk of certain cancers: According to research, coffee consumption can lower the risk of certain cancers, including liver and colorectal cancer.5. Reduced risk of heart disease: There is evidence that drinking moderate amounts of coffee can lower the risk of heart disease.Risks:1. Insomnia: Drinking too much coffee, especially late in the day, can cause insomnia.2. Increased risk of anxiety: Caffeine can increase anxiety levels in some people.3. Digestive issues: Drinking too much coffee can cause digestive problems like acid reflux, diarrhea, and ulcers.4. Increased heart rate: Caffeine can increase heart rate, which can be dangerous for people with pre-existing heart conditions.5. Addiction: Regular consumption of caffeine can lead to addiction.To maximize the health benefits and minimize the risks of drinking coffee, here are a few tips:1. Limit your caffeine intake to no more than 400 milligrams per day, or about 4 cups of coffee.2. Avoid drinking coffee late in the day, especially if you have trouble sleeping.3. Opt for organic, pesticide-free coffee to minimize exposure to toxins.4. Add cinnamon to your coffee instead of sugar to help regulate blood sugar levels.5. Drink your coffee black or with a small amount of milk to avoid consuming excessive sugar and calories.6. Consider switching to decaf coffee if you want to cut back on caffeine. 7. Avoid drinking coffee on an empty stomach as it can increase the risk of digestive problems.\\ \midrule
    I love my morning coffee, but sometimes it makes me feel a bit jittery. What can I do to avoid this? \\ \midrule
    If you love your morning coffee but find that it's making you feel jittery, here are a few tips to help you avoid this:1. Choose a lower-caffeine coffee: Look for coffee that has less caffeine, like decaf or a blend with a lower percentage of caffeine.2. Don't drink coffee on an empty stomach: Drinking coffee on an empty stomach can cause your blood sugar levels to drop, which can make you feel jittery. Try drinking your coffee with breakfast or after eating something.3. Drink water with your coffee: Stay hydrated by drinking water before or after your coffee. Drinking water will help flush out the caffeine quicker, which can help you avoid jitteriness.4. Don't drink too much coffee: Stick to one or two cups of coffee per day. Anything more than that could result in increased caffeine consumption, which can cause jitteriness.5. Switch to tea: Tea contains less caffeine than coffee, making it a good alternative if you want to avoid jitteriness. Try green tea, which has less caffeine and offers other health benefits.\\
    \bottomrule
    
    \end{tabular}
    \caption{A comparison of two dialogues with the same opening line. The former is generated by directly querying ChatGPT to output a dialogue, and the latter is generated by iteratively User-AI simulations.}
    \label{tab:compare}
\end{table*}

\end{document}